\documentclass[runningheads]{llncs}

\usepackage[T1]{fontenc}
\usepackage[utf8]{inputenc}

\usepackage{graphicx}
\usepackage{array}
\usepackage{amsmath}
\usepackage{booktabs}
\usepackage{caption}
\usepackage{multirow}
\usepackage{tabularx}
\usepackage{todonotes}
\usepackage[htt]{hyphenat}
\usepackage{tcolorbox}
\usepackage{adjustbox}
\usepackage{hyperref}

\newcolumntype{L}[1]{>{\raggedright\arraybackslash}m{#1}}
\newcolumntype{C}[1]{>{\centering\arraybackslash}m{#1}}

\begin{document}

\title{\textsc{NeoN}: A Tool for Automated Detection, Linguistic and LLM‑Driven Analysis\\of Neologisms in Polish}
\titlerunning{\textsc{NeoN}: Automated Detection and Analysis of Neologisms in Polish}

\author{Aleksandra Tomaszewska\orcidID{0000-0001-6379-3034} \and
Dariusz~Czerski\orcidID{0000-0002-3013-3483} \and
Bartosz Żuk\orcidID{0009-0008-8473-7718} \and Maciej Ogrodniczuk    \orcidID{0000-0002-3467-9424}}
\authorrunning{A. Tomaszewska, D. Czerski, B. Żuk, M. Ogrodniczuk}

\institute{Institute of Computer Science, Polish Academy of Sciences
\email{firstname.lastname@ipipan.waw.pl}}

\maketitle


\begin{abstract}
We introduce \textsc{NeoN}, a tool for detecting and analyzing Polish neologisms. Unlike traditional dictionary-based methods requiring extensive manual review, \textsc{NeoN} combines reference corpora, Polish-specific linguistic filters, an LLM-driven precision-boosting filter, and daily RSS monitoring in a multi-layered pipeline. The system uses context-aware lemmatization, frequency analysis, and orthographic normalization to extract candidate neologisms while consolidating inflectional variants. Researchers can verify candidates through an intuitive interface with visualizations and filtering controls. An integrated LLM module automatically generates definitions and categorizes neologisms by domain and sentiment. Evaluations show \textsc{NeoN} maintains high accuracy while significantly reducing manual effort, providing an accessible solution for tracking lexical innovation in Polish.\footnote{The prompt templates and a number of \textsc{NeoN} interface screenshots are available at \url{https://drive.google.com/file/d/1THipys62nlU7panPnIVdAUKzIk0QGlUx/view}}

\end{abstract}


\section{Introduction}

Neologism detection in Polish remains challenging due to the heavy reliance on dictionary-based methods and manual curation. Traditional approaches, while informative, are time-consuming and prone to human bias \cite{kerremans2018mining,klosinska2024,paryzek2008comparison}. Semi-auto\-ma\-tic systems like \textsc{Neoveille} \cite{cartier2017} and \textsc{NeoCrawler} \cite{kerremans2018mining} provide interactive interfaces for candidate review; however, their dependence on static lexicons and basic filtering mechanisms limits their ability to capture the full spectrum of emerging lexical phenomena in Polish. In addition, although recent advances in Large Language Models (LLMs) have transformed many areas of natural language processing, no existing tool has yet used LLMs for the semantic categorization and analysis of new words in Polish. In this paper, we present \textsc{NeoN} – an automated tool designed for Polish neologism detection, monitoring, and analysis.

\begin{figure}[t]
    \centering
    \includegraphics[width=0.95\textwidth]{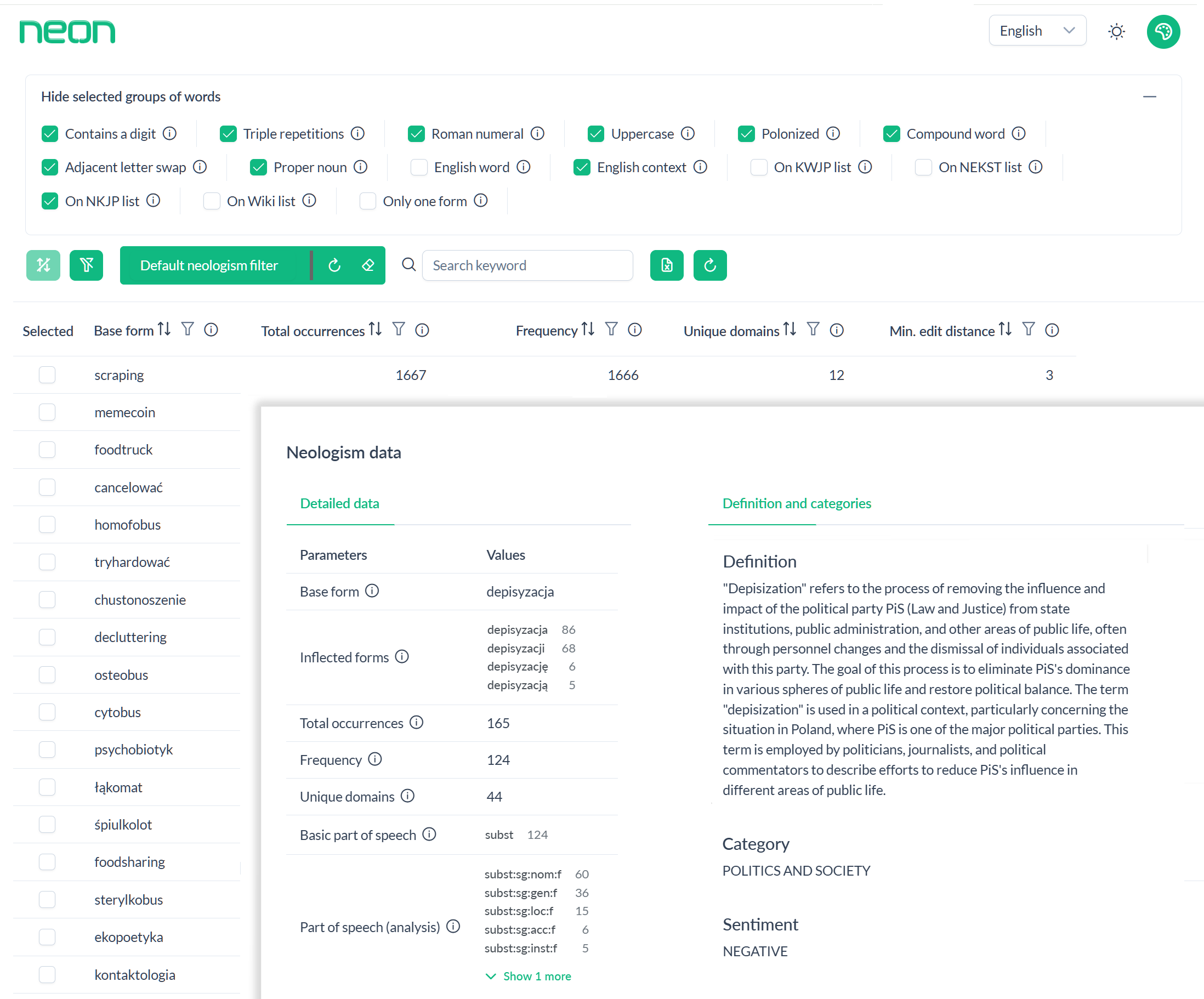}
    \caption{\textsc{NeoN}: overview of the monitoring interface.}
    \label{fig:neon-interface}
    \vspace{-5pt}
\end{figure}

Rather than relying solely on dictionary lookups, \textsc{NeoN} continuously processes RSS feeds through a multi-layered filtering pipeline. It uses four high-quality Polish corpora: the National Corpus of Polish \cite{przepiorkowski2012}, the Corpus of Contemporary Polish \cite{kieras2024}, Polish Wikipedia dump, and a web corpus NEKST \cite{czerski}. In the tool, we focus exclusively on neologisms that have entered usage after 2020 to monitor ongoing changes in the language rather than older forms. Custom filters tailored for Polish help extract candidate neologisms and consolidate variants through frequency analysis, structural constraints, context-aware lemmatization, and the final LLM-driven precision-boosting filter. \textsc{NeoN} offers a minimalistic interface accessible to users without programming expertise, displaying candidates in context with visualizations and metrics. Its integrated LLM module automatically generates definitions and categorizes neologisms by domain and sentiment, reducing manual annotation while facilitating detection and preliminary analysis. By combining corpus-based filtering with LLM-driven analysis, \textsc{NeoN} provides a scalable framework for tracking lexical innovation in Polish.

\vspace{-2pt}
\section{Related Work}
\label{related}

\vspace{-3pt}
Earlier studies, such as those by Paryzek \cite{paryzek2008comparison}, employed discriminant-based approaches to identify textual markers for flagging potential neologisms. In the Polish context, the Dictionary of Polish Neologisms, developed by the Language Observatory of the University of Warsaw \cite{klosinska2024}, serves as a recognized but manually curated resource; its reliance on diagnostic dictionaries and reference corpora underscores both its utility and limitations. Over the past two decades, semi-automated systems like \textsc{NeoCrawler} \cite{kerremans2018mining} and \textsc{Neoveille} \cite{cartier2017} have emerged, using dictionaries and rule-based filters to extract new words from sources such as websites, blogs, and press releases. Although these tools reduce manual workload, they still require expert oversight to eliminate non-neologisms, errors, and non-lexical noise, as well as to incorporate metadata (e.g., definitions and categories). Moreover, their candidate lists are typically generated through dictionary comparisons rather than by capturing neologisms as they naturally emerge, and their architectures are seldom tailored to the nuances of the Polish language.

Simultaneously, statistical and machine-learning approaches have advanced the field. For instance, Falk et al. \cite{Falk2018TheLA} proposed a framework integrating form-related, morphological, and thematic features to detect neologisms in French newspapers, while McCrae \cite{mccrae-2019} expanded on this by using pre-trained language models to differentiate compositional from non-compositional multiword expressions, thereby enhancing detection precision. More recent efforts have incorporated LLMs for additional tasks, including definition generation and translation \cite{lerner2025,zheng2024}, as well as unsupervised techniques that normalize variant forms via embedding-space mapping \cite{zalmout2019}. The dynamic nature of language on social media has also prompted research into network-based methods. Yin and Cheng \cite{7603284} applied overlapping community detection on Weibo data to track the propagation of neologisms, and Wang and Wu \cite{wang2017research} adopted similar approaches to extract new lexical items in Chinese texts. Studies of French and Italian tweets by Tarrade et al. \cite{tarrade2022detecting} and Spina et al. \cite{Spina2024}, respectively, have further demonstrated the effectiveness of these methods in capturing both the morphological evolution and semantic shifts of neologisms over time. Character-based pattern mining offers an alternative methodology, as exemplified by Lejeune and Cartier \cite{lejeune-cartier-2017-character}, who combined unsupervised data mining with supervised learning at the subword level to detect neologisms in multilingual datasets.

Another nascent line of research is whether large language models can substitute for human participants in forced elicitation tasks. One study \cite{ferrazzo2025largelanguagemodelsfuture} replicated two experiments using GPT4o-mini, whose zero-shot performance exceeded that of human informants and was further enhanced through Chain-of-Thought prompting, albeit in a limited-scale study.

\vspace{-3pt}
\section{Functionalities}

\vspace{-5pt}
Our system processes daily RSS feed data through a unified pipeline integrating candidate extraction, variant grouping, and multi-layered filtering mechanism for removing noise candidates.
All \textsc{NeoN} functionalities are integrated into an intuitive web-based interface allowing users to customize filter settings, review, adjust candidate lists, and export the resulting data in CSV format. Users can easily generate definitions and categories on demand using real-life usage examples extracted through our multi-layered pipeline. A screenshot showcasing the \textsc{NeoN} interface can be seen in Figure~\ref{fig:neon-interface}. 

Subsection~\ref{sec:form-filtering} details the various filters applied during the selection of potential neologism candidates. In Subsection~\ref{sec:form-grouping}, we describe the methods we use to address spelling variations and typographical errors. Additionally, given Polish's rich inflectional morphology, we consolidate various forms of the same lexical item through a dedicated lemmatization module. After testing several tools, we selected the best-performing \texttt{Hydra}~\cite{krasnowska2024parsing} lemmatizer. Subsections \ref{sec:definition-generation} and \ref{sec:multidimensional-categorization} describe experiments on definition generation and multidimensional categorization using \texttt{Llama-3.3-70B-Instruct} and \texttt{DeepSeek-R1} models. Ultimately, we integrated \texttt{Llama-70B} into \textsc{NeoN} as it performed comparably on definition generation and better on sentiment and domain categorization while being smaller and more cost-effective.

\vspace{-2pt}
\subsection{Form Filtering}
\label{sec:form-filtering}

\vspace{-3pt}
In neologism detection, various filters are applied to determine the likelihood of a word being new or non-standard. The key filters in \textsc{NeoN}, categorized according to their respective purposes, are:
(1) \textbf{Frequency and occurrence}: Document frequency, Term frequency, Unique domain frequency, Domain distribution; (2) \textbf{Structural constraints}: Word length constraints, Invalid character check, Presence of digits, Triple repeated characters, (3) \textbf{Lexical validation}: Common Polish corpus check, English loanword detection; (4) \textbf{Spelling and typographical errors}: Polish word matching, Edit distance with diacritics, Adjacent character swap detection; (5) \textbf{Contextual analysis}: English context detection, Name Entity heuristic, Capitalization pattern; (6) \textbf{Other}: Compound word detection, Filtering using a few-shot LLM prompt.

The first innovation in our filtering framework is incorporating language corpora alongside a dictionary as customizable references for neologism validation. Users can select up to four reference corpora based on their research needs, such as general language versus web contexts or temporal relevance. \textsc{NeoN} cross-references extracted lexemes against these corpora to exclude existing Polish words, improving detection accuracy. General reference corpora are used by default, with an option to include a web corpus for recent data. The corpora include frequency lists from the National Corpus of Polish (NKJP, up to 2010), the Corpus of Contemporary Polish (KWJP, 2011--2020), the web corpus NEKST (up to 2020), and the latest Polish Wikipedia dump.

The second innovation is employing large language models (LLMs) as the final filtering stage. We use the \texttt{Llama-3.3-70B-Instruct} model with a few-shot prompt containing three positive and three negative examples to enhance neologism identification. Recent studies show that large LLMs can outperform humans in this task \cite{zheng2024}. To our knowledge, this is the first application of LLMs as a filtering mechanism in neologism detection, advancing identification accuracy. We evaluated its effectiveness through experiments on a large corpus.

\subsubsection{Experimental Setup}
The experiments involved a corpus of 233,538 web documents (873 RSS sources) from approx. two months. The documents underwent a processing pipeline that involved language detection, main content extraction to isolate the primary textual content from web pages, and NLP analysis to process the text using the \texttt{Hydra}~\cite{krasnowska2024parsing} for NLP. Following initial filtering with a Polish language dictionary\footnote{Available at: \url{https://sjp.pl}}, we generated a set of 200,696 candidate neologisms. These candidates were then refined through multi-step filtering to distinguish neologisms from noise or established lexemes.

\vspace{-10pt}
\subsubsection{Filtering Pipeline}
The neologism filtering mechanism was implemented as an iterative sequence of filters, each designed to eliminate specific types of non-neologisms. For consistency evaluation purposes (neologisms that appeared after 2020), only selected filters were used.
(1) \textbf{Length constraints}: Words must be at least 3 characters and no longer than 20 characters; 
(2) \textbf{Numerical content}: Words must not contain digits; 
(3) \textbf{Frequency}: Words must appear in more than 5 documents; 
(4) \textbf{Case sensitivity}: Words must appear in lowercase in at least 5 occurrences; 
(5) \textbf{Proper noun exclusion}: Words must not function as proper nouns in at least 5 occurrences; 
(6) \textbf{Edit distance}: The minimum edit distance to any known word in the Polish dictionary must exceed 0.5; 
(7) \textbf{Spelling}: Words must not be diacritical variations, result from swapping adjacent letters of existing Polish dictionary words and not contain triple repetitions of the same letter; 
(8) \textbf{English dictionary check}: If a word appears in an English dictionary, it must occur in at least 5 Polish-language contexts; 
(9) \textbf{Exclusion from other dictionaries}: Words must not be present in the NKJP or KWJP dictionaries.
(10) \textbf{LLM filtering}: Filtering (we utilized the \texttt{Llama-3.3-70B-Instruct} model) based on a prepared few-shot prompt. The prompt is provided in the appendix.

\vspace{-10pt}
\subsubsection{Evaluation Dataset}
For evaluation, we used a dataset of neologisms added to the Language Observatory dataset after 2020 \cite{klosinska2024}. This timeframe aligns with our primary reference corpora. We preprocessed the dataset by removing neologisms already listed in the Polish dictionary, as these typically reflect existing words with new meanings rather than completely new lexemes. After preprocessing, our final training set included 610 neologisms. To ensure the corpus was suitable for evaluation, we expanded it by gathering the top 100 Google search results for each neologism in the training set. For a more precise evaluation, we conducted a manual review of 1,740 neologisms identified during the final filtering stage – before applying the large language model (LLM) filter – excluding those already in the Language Observatory dataset.
This method enables a more effective evaluation of our tool's precision. The results obtained through this process are presented in Table~\ref{tab:incremental-filtering} as an additional section labeled 'Including human-annotated data'. The assessment was conducted by three individuals: two annotators and one adjudicator who resolved instances of conflicting evaluations. The recall for this set remains identical to that of the test set, as the manual annotations did not change the number of detected neologisms. This process only validated the existing candidates without adding or removing any, allowing the focus to shift to precision and F1 scores based on these annotations.

\vspace{-10pt}
\subsubsection{Experimental Procedure}
The experiment was conducted iteratively, with each iteration introducing an additional filter to the pipeline. At each stage, we evaluated filtering performance using precision, recall, and F1 score, comparing filtered candidates to the testing set ground truth. The obtained results are presented in Table~\ref{tab:incremental-filtering}.

\vspace{10pt}
\begin{table}
\centering
\caption{Results of incremental filtering in the neologism detection. (*) The recall measure cannot be effectively calculated based solely on annotated data.}
\begin{tabular}{l r r c c C{1.5cm}}
\toprule
& \multicolumn{5}{c}{\textbf{Test set}} \\
\cmidrule(lr){2-6}
\textbf{Conditions} & \textbf{All}\hspace{8pt} & \textbf{Matches} & \textbf{Precision} & \textbf{Recall} & \textbf{F1} \\
\midrule
No filter & 200\,696 & 610\hspace{8pt} & 0.003 & 0.993 & 0.006 \\
+ Min Token Len & 199\,977 & 610\hspace{8pt} & 0.003 & 0.993 & 0.006 \\
+ Max Token Len & 199\,289 & 609\hspace{8pt} & 0.003 & 0.992 & 0.006 \\
+ No Digits & 186\,422 & 609\hspace{8pt} & 0.003 & 0.992 & 0.007 \\
+ Freq $\geq$ 5 & 33\,801 & 607\hspace{8pt} & 0.018 & 0.989 & 0.035 \\
+ Non-Uppercase Freq $\geq$ 5 & 5\,116 & 603\hspace{8pt} & 0.118 & 0.982 & 0.210 \\
+ Non-NE Freq $\geq$ 5 & 4\,198 & 597\hspace{8pt} & 0.142 & 0.972 & 0.248 \\
+ Min Edit Distance & 3\,130 & 552\hspace{8pt} & 0.176 & 0.899 & 0.295 \\
+ Spelling & 2\,726 & 549\hspace{8pt} & 0.201 & 0.894 & 0.329 \\
+ Non-Eng Freq $\geq$ 5 & 2\,657 & 549\hspace{8pt} & 0.207 & 0.894 & 0.336 \\
+ Not in NKJP & 1\,784 & 538\hspace{8pt} & 0.302 & 0.876 & 0.449 \\
+ Not in KWJP100 & 1\,740 & 536\hspace{8pt} & 0.308 & 0.873 & 0.455 \\
+ LLM filtering & 1\,056 & 536\hspace{8pt} & 0.508 & 0.873 & \textbf{0.642} \\
\midrule
& \multicolumn{5}{c}{\textbf{Including human-annotated data}} \\
\midrule
+ Not in KWJP100 & 1\,740 & 1\,385\hspace{8pt} & 0.796 & (*) & --- \\
+ LLM filtering & 1\,056 & 968\hspace{8pt} & \textbf{0.917} & 0.699 & \textbf{0.793} \\
\bottomrule
\end{tabular}
\label{tab:incremental-filtering}
\end{table}

\subsubsection{Summary of Results}
The neologism detection pipeline, integrating rule-based filters and a large language model (LLM), effectively identified new Polish lexemes in a noisy web corpus, achieving F1 scores of 0.642 on the test set and 0.793 on the annotated set. At the final filtering stage (i.e., before applying the LLM filter), recall cannot be computed because the process does not retain information about false negatives. In earlier stages, only the candidates that passed the filter are tracked, so any items mistakenly removed (false negatives) are lost, making it impossible to determine recall accurately. Starting with 200,696 candidates, the pipeline reduced this to 1,056 highly probable neologisms, with precision rising to 0.508 and recall settling at 0.873. Rule-based filters drastically reduce non-neologistic candidates with minimal recall loss, while the LLM filter increases precision from 0.308 to 0.508 (test set) and from 0.796 to 0.917 (annotated set) by leveraging contextual and semantic cues. This demonstrates that modern LLM models can substantially enhance the neologism detection process. Although some recall was sacrificed (settling at 0.699) for significant precision gains, this trade-off proved valuable for applications where false positives are costly, such as in linguistic studies or when tracking emerging terminology. Each filter contributed to a stepwise improvement, making this approach highly effective.


\subsection{Form Grouping}
\label{sec:form-grouping}

\textsc{NeoN} detects neologisms' alternative spellings, inflectional forms, and syntactic variants, including multi-word forms (e.g., \textit{tusko-bus}, \textit{tuskobus}, \textit{tusko bus}). In post-processing, it identifies related forms like hyphenated or spaced variants and aggregates their frequencies. Lemmatization groups inflectional forms under a base form, which is essential given Polish's rich morphology and the novelty of neologisms. We evaluated four tools: \texttt{Stanza} \cite{qi2020stanza} and \texttt{spaCy} \cite{spacy2020} (general NLP toolkits), \texttt{Hydra} \cite{krasnowska2024parsing} (Polish-specific), LLMs \texttt{GPT4o} \cite{openai2024gpt4technicalreport} and \texttt{DeepSeek-R1} \cite{deepseekai2025deepseekr1incentivizingreasoningcapability} with custom prompts. Using a dataset from \textit{Language Observatory} website\footnote{\url{https://obserwatoriumjezykowe.uw.edu.pl}}, which contains the base forms of neologisms, their inflectional variants, definitions, metadata, and example sentences illustrating their contextual usage, filtered to 978 neologisms with $\geq$3 inflectional forms each (3,659 total), we assessed their effectiveness (see Table~\ref{tab:neologism_stats}). This dataset provides a robust foundation for evaluating the effectiveness of lemmatization tools in handling Polish neologisms.

\begin{table}
\centering
\caption{Statistics of the neologisms dataset.}
\label{tab:neologism_stats}
\begin{tabular}{>{\raggedright\arraybackslash}p{7cm}c}
\toprule
\textbf{Statistic} & \textbf{Value} \\
\midrule
Total number of neologisms & 1\,790 \\
Average number of forms per neologism & 3.13 \\
Average number of examples per neologism & 3.75 \\
Number of neologisms with a definition & 1\,520 \\
Number of neologisms with metadata & 1\,787 \\
Number of neologisms with at least 3 forms & 1\,248 \\
Number of neologisms with no forms & 274 \\
Number of neologisms with no examples & 274 \\
\bottomrule
\end{tabular}
\end{table}

While traditional accuracy metrics focus on individual form correctness, they fail to capture consistency across inflectional groups – critical for applications like neologism detection. We propose two group-based metrics to address this gap, both calculated as ratios over the total number of neologism groups \( G \):

\begin{description}
    \item[Group Accuracy (Consistency)] \( A_{\text{gr}} = \frac{S}{G} \), where \( S \) is the number of groups with all inflected forms sharing the same lemma (correct or not). This ensures consistent frequency counts, e.g., ''metaverse'' and ''metaverses'' both mapped to ''metavar''.
    \item[Strict Group Accuracy] \( A_{\text{strict}} = \frac{K}{G} \), where \( K \) is the number of groups with all forms correctly mapped to the base form. This is critical for precise linguistic studies, e.g., tracking ''cancel culture''.
\end{description}

\subsubsection{Experimental Setup and Objectives}
We ran two experiments to assess the lemmatization of neologisms: one where models lemmatize isolated single words (e.g., "\textit{NFTs}" → "\textit{NFT}") without context to test inherent consistency, and another where models lemmatize within sentence contexts (e.g., "\textit{They NFTs gained popularity}" → "\textit{NFT}"), leveraging context to resolve ambiguity while potentially reducing group consistency.





\begin{table}[ht]
\centering
\caption{Neologism lemmatization results.}
\label{tab:lemmatization_context_nocontext_results}
\setlength{\tabcolsep}{0.15cm}
\begin{tabular}{llcp{0.01cm}cc}
\toprule
\multirow{2}{*}{\textbf{Experiment}} & \multirow{2}{*}{\textbf{Model}} & \multirow{2}{*}{\textbf{Accuracy}} && \textbf{Group} & \textbf{Strict Group} \\
& & && \textbf{Accuracy} & \textbf{Accuracy} \\
\midrule
\multirow{5}{*}{\begin{tabular}[c]{@{}l@{}}Without \\ context\end{tabular}} 
& \texttt{SpaCy}            & 50.18\%  && 14.52\%  & 13.50\%  \\
& \texttt{Stanza}           & 73.41\%  && \textbf{53.58\%} & \textbf{50.41\%}  \\
& \texttt{Hydra}            & 72.01\%  && 49.08\%  & 46.22\%  \\
& \texttt{GPT4o}       & 72.81\%  && 53.07\%  & 49.90\%  \\
& \texttt{DeepSeek-R1}      & \textbf{75.13\%} && 51.53\%  & 49.80\%  \\
\midrule
\multirow{5}{*}{\begin{tabular}[c]{@{}l@{}}With \\ context\end{tabular}} 
& \texttt{SpaCy}            & 52.94\%  && 16.26\%  & 15.44\%  \\
& \texttt{Stanza}           & 73.35\%  && 51.94\% & 48.77\%  \\
& \texttt{Hydra}            & \textbf{79.31\%} && 62.47\% & \textbf{60.22\%} \\
& \texttt{GPT4o}       & 78.57\%  && \textbf{62.99\%}  & 59.41\%  \\
& \texttt{DeepSeek-R1}      & 77.51\%  && 57.16\%  & 55.32\%  \\
\bottomrule
\end{tabular}
\end{table}

\subsubsection{Summary of Results}
The experiments on neologism lemmatization reveal significant performance variations across models and conditions. Basic NLP tools like SpaCy exhibit poor results, with accuracies of 50.18\% (without context) and 52.94\% (with context), while Stanza performs better, achieving 73.41\% and 73.35\%, respectively, which is surprising since context typically improves lemmatization. This slight decline suggests potential issues with Stanza’s ability to leverage context for Polish neologisms, warranting further investigation to determine its cause and implications. The Hydra system, tailored for Polish, excels with context, reaching the highest accuracy of 79.31\% and strict group accuracy of 60.22\%, outperforming other models. Large language models (LLMs), such as \texttt{GPT4o} and \texttt{DeepSeek-R1}, demonstrate strong potential, particularly in context-free scenarios (e.g., \texttt{DeepSeek-R1} at 75.13\% accuracy), and remain competitive with context (78.57\% and 77.51\%, respectively). These findings highlight the inadequacy of basic NLP tools for neologisms, the effectiveness of Hydra in leveraging context, and the promising capabilities of LLMs for rare word lemmatization. Future research should focus on fine-tuning a specialized LLM that integrates Hydra's contextual strengths with LLMs' robustness, improving the lemmatization of rare Polish words.



\subsection{Definition Generation}
\label{sec:definition-generation}
We conducted experiments to evaluate Large Language Models' capability to automatically generate definitions for neologisms, focusing specifically on the most recent lexemes. 
We only selected neologisms registered in 2024 from the University of Warsaw's Language Observatory dictionary. For each neologism, we obtained definitions and usage examples from the Language Observatory's website. We filtered out entries with fewer than 5 usage examples, resulting in a final test dataset of 81 neologisms.
Our experiments utilized two models: \texttt{Llama-3.3-70B-Instruct} \cite{grattafiori2024llama3herdmodels}, with a knowledge cutoff date of December 2023 and \texttt{DeepSeek-R1} \cite{deepseekai2025deepseekr1incentivizingreasoningcapability}, which had no known cutoff date as of February 28, 2025. We chose \texttt{DeepSeek-R1} particularly to compare newer reasoning-focused models against traditional LLMs like \texttt{Llama-70B}.


\subsubsection{Evaluation Protocol}
\label{sec:llm-evaluation-protocol}
We test three prompting setups: (1) the 0-shot setup where we do not provide any examples of neologism usage, (2) 3-shot, and (3) 5-shot where we provide 3 and 5 examples of their usage, respectively. 
For all the experiments, we sampled the models using the recommended temperature of $0.6$ and top-$p$ value of $0.95$\footnote{For details see \url{https://huggingface.co/deepseek-ai/DeepSeek-R1} and \\ \url{https://huggingface.co/meta-llama/Llama-3.3-70B-Instruct}.}.
We evaluated the generated definitions using the ''LLM-as-a-judge'' approach \cite{zheng2023judgingllmasajudgemtbenchchatbot}, which employs LLMs to score, rank, or select from candidate options. For our experiments, we used the \texttt{GPT4o} model \cite{openai2024gpt4technicalreport} (knowledge cutoff: October 2023) as the judge, performing both pointwise and pairwise evaluations: (1) \textbf{Pointwise Evaluation}: the judging LLM compared the generated definition against a human-made reference, outputting \texttt{CORRECT} or \texttt{INCORRECT}. This evaluation setup largely follows \cite{zheng2024} and focuses exclusively on the correctness of the definition. (2) \textbf{Pairwise Evaluation}: the judging LLM compared model-generated and human-made definitions, outputting \texttt{WIN} (first definition better), \texttt{DRAW} (equal quality), or \texttt{LOSE} (first definition worse). While also taking into account correctness, this evaluation setup relies on other criteria like style, readability, and conciseness.
To increase quality, we included all five usage examples in the prompt. To mitigate known positional bias in LLM judges \cite{qin-etal-2024-large,li2023split}, we randomly shuffled the order of competing definitions in pairwise evaluations. 

\subsubsection{Summary of Results}
\begin{figure}[h]
\centering
\begin{minipage}{0.52\textwidth}
\centering
\includegraphics[width=\textwidth]{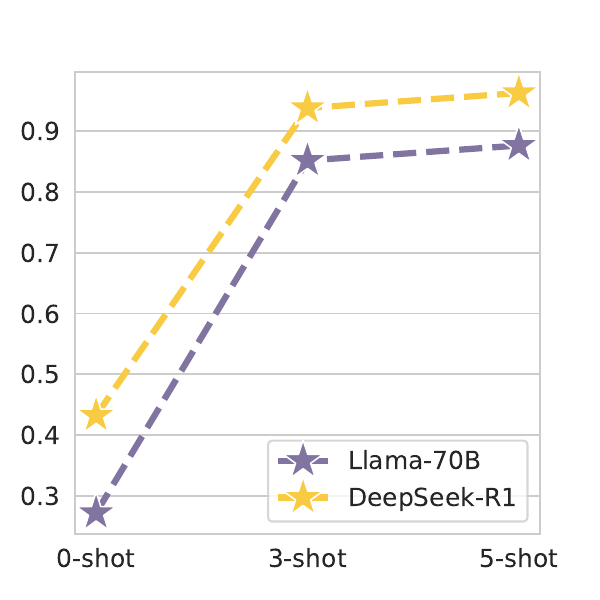} 
\caption{Accuracy of \texttt{DeepSeek-R1} and \texttt{Llama-70B} in pointwise evaluation across three prompting setups.}
\label{fig:correctness-winrate}
\end{minipage}\hfill
\begin{minipage}{0.44\textwidth}
\centering
\captionof{table}{Detailed results for pointwise evaluation of \texttt{DeepSeek-R1} and \texttt{Llama-70B} across three prompting setups.}
\label{tab:correctness-winrate}
\begin{tabular}{l c c}
\toprule
& \multicolumn{2}{c}{\textbf{Verdict}} 
\tabularnewline \cmidrule(lr){2-3}
& Correct & Incorrect \\
\midrule
\texttt{Llama-70B} 0-shot & 22 & 59 \\
\texttt{Llama-70B} 3-shot & 69 & 12 \\
\texttt{Llama-70B} 5-shot & 71 & 10 \\
\texttt{DeepSeek-R1} 0-shot & 35 & 46 \\
\texttt{DeepSeek-R1} 3-shot & 76 & 5 \\
\texttt{DeepSeek-R1} 5-shot & 78 & 3 \\
\bottomrule
\end{tabular}
\end{minipage}
\end{figure}

Figure~\ref{fig:correctness-winrate} shows the accuracy of both models in the pointwise evaluation. Performance improved monotonically with additional usage examples. \texttt{DeepSeek-R1} outperformed \texttt{Llama-70B} across all setups, achieving the maximum 96\% accuracy in the 5-shot setup compared to \texttt{Llama-70B}'s 88\%. Table~\ref{tab:correctness-winrate} presents detailed results for each experimental setup.

\begin{figure}[h]
\centering
\begin{minipage}{0.48\textwidth}
\centering
\includegraphics[width=\textwidth]{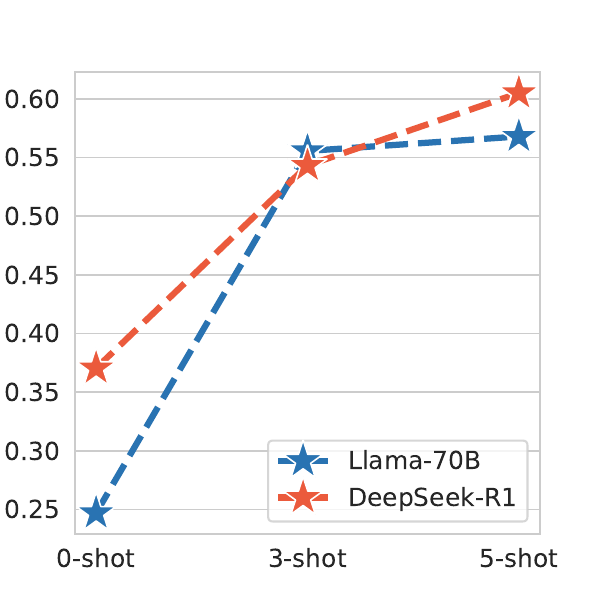} 
\caption{Win rate of \texttt{DeepSeek-R1} and \texttt{Llama-70B} in pairwise evaluation against human-made definition across three prompting setups.}
\label{fig:human-comparison-winrate}
\end{minipage}\hfill
\begin{minipage}{0.48\textwidth}
\centering
\captionof{table}{Detailed results for pairwise evaluation of \texttt{DeepSeek-R1} and \texttt{Llama-70B} against human-made definition across three prompting setups.}
\label{tab:human-comparison-winrate}
\begin{tabular}{l c c c}
\toprule
& \multicolumn{3}{c}{\textbf{Verdict}} 
\tabularnewline \cmidrule(lr){2-4}
& Win & Draw & Lose \\
\midrule
\texttt{Llama-70B} 0-shot & 20 & 1 & 60 \\
\texttt{Llama-70B} 3-shot & 45 & 4 & 32 \\
\texttt{Llama-70B} 5-shot & 46 & 5 & 30 \\
\texttt{DeepSeek-R1} 0-shot & 30 & 4 & 47 \\
\texttt{DeepSeek-R1} 3-shot & 44 & 4 & 33 \\
\texttt{DeepSeek-R1} 5-shot & 49 & 9 & 23 \\
\bottomrule
\end{tabular}
\end{minipage}
\end{figure}

Figure~\ref{fig:human-comparison-winrate} shows the win rate of model-generated definitions against human-made ones in the pairwise evaluation. As with pointwise evaluation, performance improved with additional usage examples, highlighting their importance. Interestingly, \texttt{Llama-70B} slightly outperformed \texttt{DeepSeek-R1} in the 3-shot setup (55\% vs. 54\% win rate). In the 5-shot setup, \texttt{DeepSeek-R1} led by a narrow margin of 4\%, showing that \texttt{Llama-70B} remains competitive despite having nearly 10x fewer parameters (70B vs. 671B). Detailed results showing wins, draws and losses for each experimental setup are available in Table~\ref{tab:human-comparison-winrate}. 

\subsubsection{Meta Evaluation}
Upon manual inspection, we conducted a meta evaluation verifying \texttt{GPT4o}'s effectiveness as an LLM judge. Using three human annotators we evaluate the generated neologism definitions. We focus only on the 5-shot setup as it produces the best results across both models. The evaluation followed our previously described protocol. To hide the origin of the definitions and minimize potential bias the evaluation was separated into two phases done in a particular order: (1) \textbf{Pairwise Meta Evaluation}: Annotators compared randomly shuffled definitions without knowing their origin. (2) \textbf{Pointwise Meta Evaluation}: Annotators evaluated definition correctness using human-made ones as reference. For this phase, they were informed which definitions were model-generated.

\begin{figure}[!h]
\centering
\includegraphics[width=\linewidth]{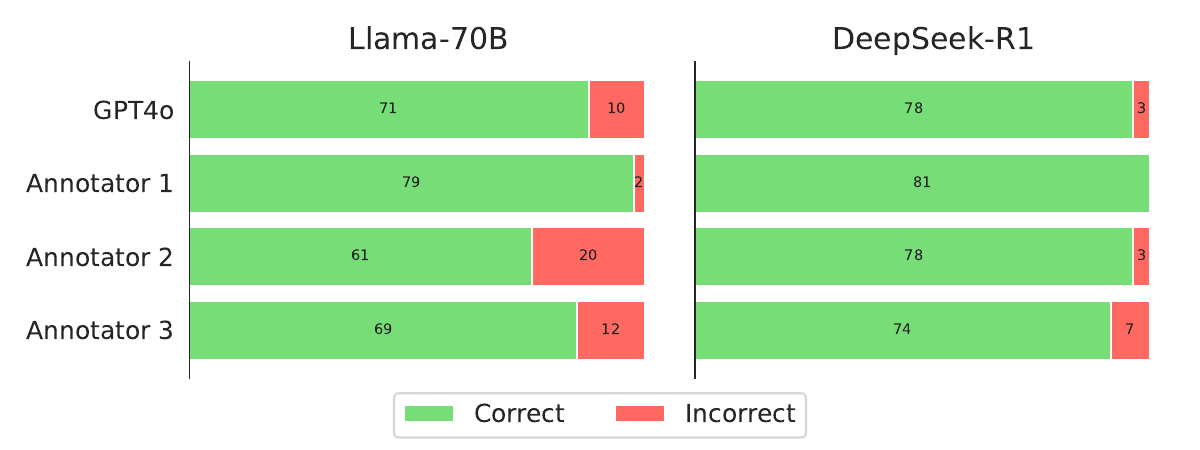}
\caption{Results of pointwise meta evaluation shown for 3 human annotators and \texttt{GPT4o} (LLM judge) across two judged models: \texttt{Llama-70B} and \texttt{DeepSeek-R1}.}
\label{fig:correctness-meta}
\end{figure}

The results of the meta evaluation for pointwise evaluation are presented in Figure~\ref{fig:correctness-meta}. Human annotators showed high agreement with \texttt{GPT4o}'s judgments, consistently rating \texttt{Llama-70B} lower than \texttt{DeepSeek-R1}, which aligns with the results in Figure~\ref{fig:correctness-winrate}. Annotators varied in their strictness: Annotators 2 and 3 marked more definitions as incorrect compared to \texttt{GPT4o}, while Annotator 1 was more lenient, marking only two \texttt{Llama-70B} definitions as incorrect and none for \texttt{DeepSeek-R1}.

\begin{figure}[h]
\centering
\includegraphics[width=\linewidth]{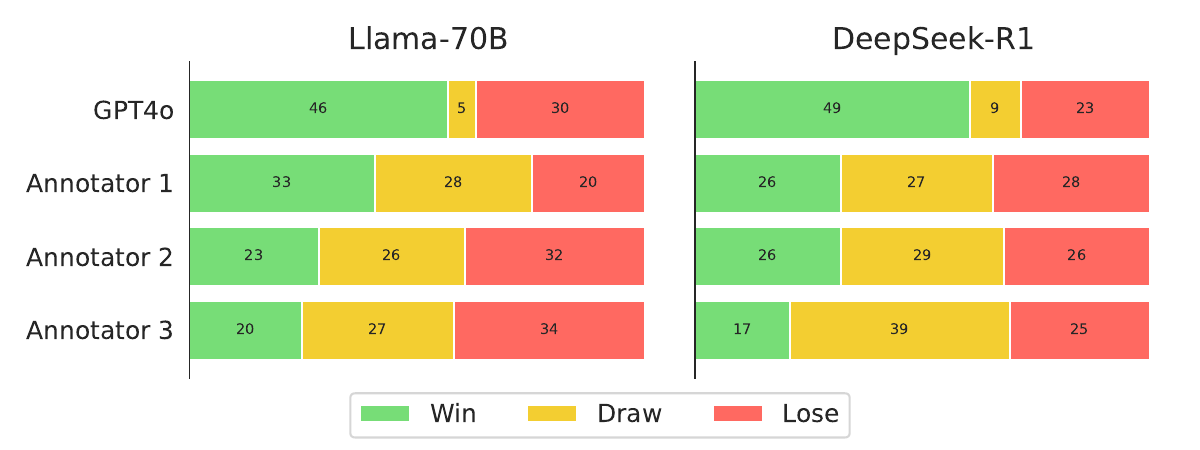}
\caption{Results of pairwise meta evaluation, shown for 3 human annotators and \texttt{GPT4o} (LLM judge) across two judged models: \texttt{Llama-70B} and \texttt{DeepSeek-R1}.}
\label{fig:human-comparison-meta}
\end{figure}

Figure~\ref{fig:human-comparison-meta} shows the pairwise meta evaluation results. \texttt{GPT4o} demonstrates a clear bias toward model-generated definitions, with human annotators' win rates being consistently lower than LLM evaluations. Notably, Annotators 1 and 3 assigned more wins to \texttt{Llama-70B} than to \texttt{DeepSeek-R1}. Indicating that while \texttt{DeepSeek-R1} performs better in terms of correctness, \texttt{Llama-70B} seems to match it in other criteria such as style and conciseness that are naturally taken into account during pairwise evaluation. This finding aligns with results shown in Figure~\ref{fig:human-comparison-winrate}, where \texttt{Llama-70B} competes well and even outperforms \texttt{DeepSeek-R1}'s win rates in the 3-shot setup.

\subsection{Multidimensional Categorization}
\label{sec:multidimensional-categorization}

We evaluate \texttt{Llama-70B} and \texttt{DeepSeek-R1} on neologism categorization across sentiment and domain. The LLM evaluates each neologism's sentiment as positive, neutral, or negative. For example, \textit{unfriendować} (''to unfriend'') typically carries a negative sentiment. Additionally, the LLM classifies each neologism into one of six domains: (1) \textsc{Technology and Science}: lexemes emerging from scientific and technological advances, such as \textit{halucynować} (''to hallucinate'') when describing AI-generated plausible but fabricated information; (2) \textsc{Culture and Entertainment}: originating in art, media, literature, and popular culture, (3) \textsc{Social Life and Relationships}: reflecting evolving social interactions and communication patterns, (4) \textsc{Economy and Business}: driven by economic trends and business practices, (5) \textsc{Ecology and Environment}: associated with environmental issues and sustainability, (6) \textsc{Politics and Society}: reflecting political changes and social movements.

\subsubsection{Evaluation Protocol}
We test three prompting setups: (1) \textsc{Examples} setup which includes 5 examples of neologism usage in the prompt, (2) \textsc{Definition} setup which provides an automatically generated definition for the given neologism and (3) \textsc{Both} which combines two previous setups. We manually categorize the 81 neologisms from test set introduced in Section~\ref{sec:definition-generation}. For evaluation, we use two metrics: Accuracy and Macro-F1 which is the average F1 score across all classes, giving equal weight to each class regardless of its size in the imbalanced dataset.


\subsubsection{Summary of Results}
Table~\ref{tab:categorization} presents the multidimensional categorization results for each model and prompt setup. \texttt{Llama-70B} outperforms \texttt{DeepSeek-R1} across all prompting configurations, despite being a significantly smaller model. This may suggest that \texttt{DeepSeek-R1}'s reasoning-focused reinforcement learning phase may impair its performance on more basic tasks like classification. The \textit{Examples} setup achieves the highest Macro-F1 scores for both dimensions, reaching $0.7375$ for sentiment and $0.4341$ for domain categorization.

\begin{table}[h]
\centering
\caption{Multidimensional categorization results for \texttt{Llama-70B} and \texttt{DeepSeek-R1} across three prompt setups.}
\label{tab:categorization}
\begin{tabular}{l c c c c}
\toprule
& \multicolumn{2}{c}{\textbf{Sentiment}} & \multicolumn{2}{c}{\textbf{Domain}} 
\tabularnewline \cmidrule(lr){2-3} \cmidrule(lr){4-5}
& \textbf{Accuracy} & \textbf{Macro F1} & \textbf{Accuracy} & \textbf{Macro F1} \\
\midrule
\texttt{Llama-70B} + Examples & \textbf{0.7901} & \textbf{0.7375} & 0.6420 & \textbf{0.4341} \\
\texttt{Llama-70B} + Defintion & 0.7407 & 0.6970 & 0.6914 & 0.4063  \\
\texttt{Llama-70B} + Both & 0.7284 & 0.6750 & \textbf{0.7160} & 0.4105  \\
\texttt{DeepSeek-R1} + Examples & 0.7037 & 0.6424 & 0.7037 & 0.4221  \\
\texttt{DeepSeek-R1} + Defintion & 0.7407 & 0.6815 & 0.6420 & 0.4082  \\
\texttt{DeepSeek-R1} + Both & 0.6914 & 0.6287 & 0.6667 & 0.4040  \\
\bottomrule
\end{tabular}
\end{table}



\section{Conclusions and Future Work}

We present \textsc{NeoN}, our novel tool for the automated detection and analysis of Polish neologisms. Our contributions encompass an expert-driven design process informed by an overview of existing applications, leading to a comprehensive system that monitors and ranks candidate neologisms from RSS feeds by cross-checking them against a dictionary and four high-quality corpora. \textsc{NeoN} further analyzes frequency trends through graphical displays, performs thematic categorization, and generates automatic linguistic descriptions. By integrating LLMs, the system boosts the final stage of candidate filtering, generates definitions, and performs multidimensional categorization based on real-life contexts derived from the RSS data. Experiments with three linguists confirm that \textsc{NeoN} reduces manual intervention and maintains high detection accuracy and annotation quality, thereby providing a scalable, user-friendly resource for researchers without programming expertise. 

Looking ahead, we plan to enable users to upload their own corpora, reducing our reliance on RSS feeds. This will allow us to incorporate a broader range of genres, such as social media and blogs, to capture neologisms in real time and improve the representativeness of our linguistic analyses. At the same time, we aim to integrate advanced deep learning techniques to refine morphological disambiguation and semantic analysis. By incorporating real-time expert feedback and expanding filtering options based on user input, we expect to develop a more dynamic and responsive system. Additionally, we plan to use open LLMs to introduce new features. Our future work will focus on developing standardized benchmark datasets and evaluation metrics specifically for neologism detection and classification. Building on our promising initial results, we will fine-tune LLMs for this task by providing both manual and semi-automatic annotated instructions, generating synthetic examples to improve base form detection, and implementing LLM-based detection techniques. We also see potential in training smaller, specialized models that can be deployed more efficiently without compromising performance.



\clearpage
\bibliographystyle{splncs04}
\bibliography{main}

\end{document}